\pdfoutput=1

\documentclass[11pt]{article}
\usepackage{float}

\usepackage[final]{acl}

\usepackage{times}
\usepackage{latexsym}
\usepackage{subcaption}

\usepackage[T1]{fontenc}

\usepackage[utf8]{inputenc}

\usepackage{microtype}

\usepackage{inconsolata}

\usepackage{graphicx}

\usepackage{lipsum}
\usepackage{booktabs}
\usepackage{multirow}
\usepackage{graphicx}
\usepackage{tabularx}
\usepackage[export]{adjustbox}
\usepackage{amsmath}
\usepackage{amssymb}  
\usepackage{enumitem}
\usepackage{float}
\usepackage{placeins}
\usepackage{afterpage}
\usepackage{tcolorbox}
\usepackage{colortbl}
\usepackage{array, xcolor}
\usepackage{tabularray}

\title{From Prompting to Preference Optimization: A Comparative Study of LLM-based Automated Essay Scoring}

\author{
    Minh Hoang Nguyen*, Vu Hoang Pham, Xuan Thanh Huynh, Phuc Hong Mai, \\
    \textbf{Vinh The Nguyen, Quang Nhut Huynh, Huy Tien Nguyen, Tung Le} \\
    Faculty of Information Technology, University of Science, Ho Chi Minh City, Vietnam \\
    Vietnam National University, Ho Chi Minh City, Vietnam \\
    \texttt{\{\href{mailto:24C15049@student.hcmus.edu.vn}{\color{black}{24C15049}},
    \href{mailto:24C11036@student.hcmus.edu.vn}{\color{black}{24C11036}},
    \href{mailto:24C11072@student.hcmus.edu.vn}{\color{black}{24C11072}},
    \href{mailto:24C15018@student.hcmus.edu.vn}{\color{black}{24C15018}},
    \href{mailto:24C11035@student.hcmus.edu.vn}{\color{black}{24C11035}},
    \href{mailto:24C15021@student.hcmus.edu.vn}{\color{black}{24C15021}}\}@student.hcmus.edu.vn} \\
    \texttt{\{\href{mailto:ntienhuy@fit.hcmus.edu.vn}{\color{black}{ntienhuy}},
    \href{mailto:lttung@fit.hcmus.edu.vn}{\color{black}{lttung}}\}
    @fit.hcmus.edu.vn}
}

\begin{document}
\maketitle

\begin{abstract}
Large language models (LLMs) have recently reshaped Automated Essay Scoring (AES), yet prior studies typically examine individual techniques in isolation, limiting understanding of their relative merits for English as a Second Language (L2) writing. To bridge this gap, we  presents a comprehensive comparison of major LLM-based AES paradigms on IELTS Writing Task~2. On this unified benchmark, we evaluate four approaches: (i) encoder-based classification fine-tuning, (ii) zero- and few-shot prompting, (iii) instruction tuning and Retrieval-Augmented Generation (RAG), and (iv) Supervised Fine-Tuning combined with Direct Preference Optimization (DPO) and RAG. Our results reveal clear accuracy–cost–robustness trade-offs across methods, the best configuration, integrating k-SFT and RAG, achieves the strongest overall results with F1-Score 93\%. This study offers the first unified empirical comparison of modern LLM-based AES strategies for English L2, promising potential in auto-grading writing tasks. Code is public at \url{https://github.com/MinhNguyenDS/LLM_AES-EnL2}.
\end{abstract}

\section{Introduction}
Automated Essay Scoring (AES) aims to automatically evaluate written essays by assigning proficiency scores that are consistent with human raters.
The task has attracted long-standing interest in natural language processing (NLP) due to its potential to reduce grading cost, improve scoring consistency, and provide timely feedback to learners \cite{zawacki2019systematic,lagakis2021automated}. In second language (L2) writing assessment, such as IELTS or TOEFL, AES systems are expected not only to predict an overall proficiency score but also to reflect complex rubric-based criteria, including task response, coherence, lexical usage, and grammatical accuracy \cite{storch2009impact,oh2020second,li2022assessing}.

Over the past decades, AES systems have evolved from feature-engineering approaches utilizing surface-level metrics \citep{chen2013automated} to deep neural networks, such as Long Short-Term Memory (LSTM) networks \citep{taghipour2016neural,nguyen2024rrs} and pre-trained transformers like BERT and RoBERTa \citep{dong2017attention,wu2022beyond,nguyen2025co}. While discriminative models excel at capturing coherence, they often treat scoring as a simple regression or classification task, potentially overlooking the nuanced reasoning required to justify a score against complex rubrics \citep{ramesh2022automated}.

\begin{figure}[!t]
    \centering
    \includegraphics[width=\columnwidth]{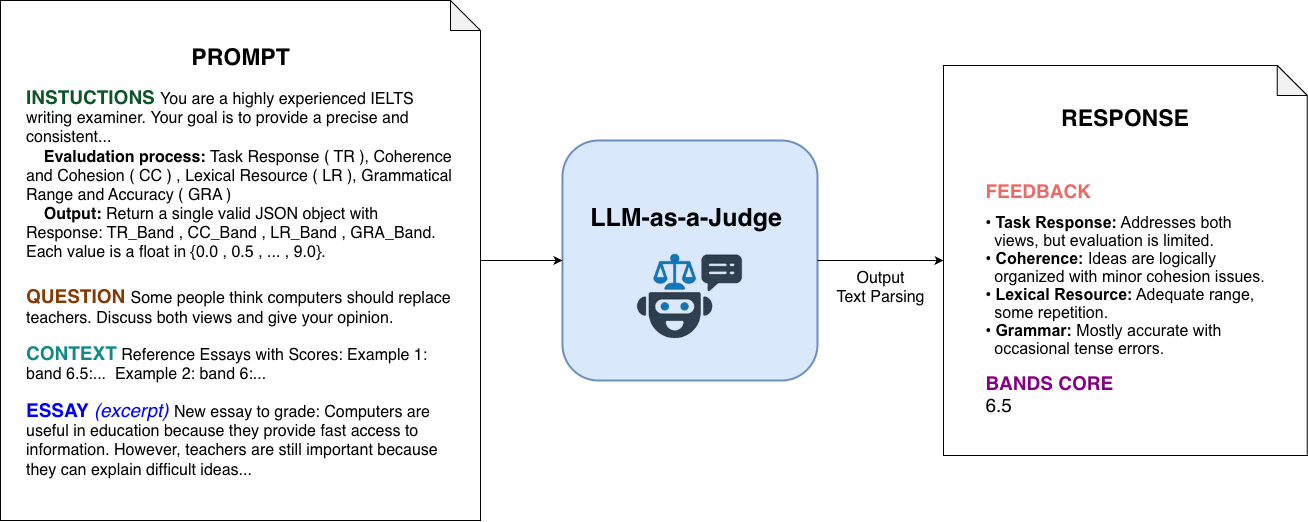}
    \caption{ A framework of LLM-based Automated Essay Scoring. Given an IELTS Writing Task~2 prompt (e.g., ``Some people think computers should replace teachers'') and a student essay response, the LLM acts as an examiner to produce (i) an overall band score (e.g., 6.5) and (ii) rubric-aligned feedback across Task Response, Coherence and Cohesion, Lexical Resource, and Grammatical Range and Accuracy.}
    \label{fig:intro}
\end{figure}

The advent of Large Language Models (LLMs) has ushered in a new paradigm for AES (Figure~\ref{fig:intro}) \citep{pack2024large}. Transformer-based models pretrained on massive corpora enabling new paradigms such as in-context learning and instruction following \citep{brown2020language}. Generative models like GPT-4 and Llama-3 have demonstrated remarkable capabilities in zero-shot and few-shot inference, offering not just scores but potential qualitative feedback \citep{kasneci2023chatgpt, roumeliotis2023chatgpt}. Recent studies have explored the potential of LLMs, including LLaMA, and GPT-4-class models, for AES and L2 writing assessment, reporting promising correlations with human scores on benchmarks such as TOEFL and IELTS \citep{mizumoto2023exploring,mansour2024can}. For instance, \citep{mizumoto2023exploring} utilized ChatGPT for TOEFL essays, and recent work has compared GPT-4o against fine-tuned RoBERTa models for IELTS \citep{qiu2024large}. These results suggest that LLMs may overcome some limitations of earlier neural AES, especially in feedback capability due to their stronger language understanding, and reasoning. 

However, existing work on LLM-based AES remains fragmented. Most studies analyze these techniques in isolation—focusing solely on prompt engineering \citep{mansour2024can}, regression fine-tuning \citep{sun2024automatic}, or pairwise ranking \citep{yang2020enhancing}—without systematically comparing how these disparate paradigms interact. Furthermore, while techniques such as RAG~\citep{lewis2020retrieval} and Reinforcement Learning from preferences have shown effectiveness in other NLP tasks \citep{jiang2024survey}, their systematic impact on AES, especially for English L2 writing.

In this paper, we address this gap by presenting a systematic and comprehensive comparison of LLM-based AES strategies applied to IELTS Writing Task 2.  Using IELTS Writing Task~2 as a common benchmark, we evaluate four representative approaches: (1) Discriminative Fine-Tuning using encoder-based models; (2) In-Context Learning via zero-shot and few-shot prompting; (3) Instruction Tuning combined with RAG to ground scoring in rubric-specific examples; and (4) Supervised Fine-Tuning (SFT) aligned via Reinforcement Learning (DPO). Our primary contributions are as follows:
\begin{itemize}
    \item We introduce the first unified benchmark that systematically compares four major LLM adaptation paradigms for English L2 automated essay scoring.
    \item We design and evaluate a criterion-aware AES framework that combines instruction tuning, retrieval-augmented generation, and preference optimization to align LLM predictions with IELTS scoring rubrics.
    \item Through extensive quantitative and qualitative analysis, we reveal clear accuracy--cost--robustness trade-offs across paradigms, offering practical guidance for deploying LLM-based AES systems.
\end{itemize}
\section{Related Work}
\subsection{Automated Essay Scoring (AES)}
Automated Essay Scoring (AES) aims to automatically assign scores to written essays in a way that approximates human raters, thereby enabling scalable and cost-effective assessment in educational settings \citep{lagakis2021automated}. AES has been widely deployed in large-scale language proficiency examinations such as TOEFL, GRE, and IELTS, as well as in classroom-oriented writing platforms and online learning systems for English as a Second or Foreign Language (EFL/ESL) learners \citep{banno2024can, mansour2024can}. Early work in AES primarily targeted holistic scores \citep{attali2006automated}, but subsequent research has increasingly emphasized analytic assessment, providing separate scores for dimensions such as grammar, vocabulary, coherence, and task achievement, which are especially important in L2 pedagogy \cite{stahl2024exploring}. Recent surveys underscore AES's progression, \cite{li2024automated} highlight milestones from heuristics to LLMs, noting persistent challenges in cross-prompt generalization and multi-trait scoring. Similarly, \cite{ramesh2022automated} systematically review AES techniques, emphasizing the shift toward explainable AI for educational feedback.

\begin{table*}[t]
\centering
\small
\setlength{\tabcolsep}{5pt}
\begin{tabular}{
>{\raggedright\arraybackslash}p{2.2cm}
>{\raggedright\arraybackslash}p{4.3cm}
>{\raggedright\arraybackslash}p{4cm}
>{\raggedright\arraybackslash}p{4.1cm}
}
\toprule
\textbf{Era} &
\textbf{Representative Systems} &
\textbf{Strengths} &
\textbf{Limitations} \\
\midrule

1960s--2010s \par (Traditional AES)
&
e-rater \citep{attali2006automated}
&
Interpretable, low computational cost, effective with small datasets
&
Limited semantic understanding, heavy feature engineering, weak cross-prompt generalization \\

\midrule

2010s--2020s \par (Neural AES)
&
LSTM-AES \citep{taghipour2016neural}; BERT-based AES \citep{yang2020enhancing}; multi-scale models \citep{wang2022use}
&
Better semantic modeling and improved correlation with human scores
&
Requires large labeled datasets, limited explainability, prone to prompt overfitting \\

\midrule

2020s+ \par (LLM-based, Prompting)
&
GPT-4 prompting \citep{mizumoto2023exploring}; LCES \citep{shibatamiyamura2025lces}
&
No training required, strong reasoning ability, flexible rubric adaptation
&
High inference cost, sensitivity to prompt design, and calibration issues \\

\midrule

2020s+ \par (LLM-based, Fine-tuning)
&
Fine-tuned RoBERTa \citep{qiu2024large}; instruction-tuned LLMs \citep{sun2024automatic}
&
Stable performance and improved alignment with scoring rubrics
&
Expensive annotation and isolated evaluation of single techniques \\

\midrule

2020s+ \par (LLM-based, Hybrid)
&
Hybrid LLM-based AES systems \citep{li2024automated, wang2025llms}
&
Improved robustness, reduced hallucination, and higher feedback quality
&
Fragmented evaluation and lack of unified benchmarks \\

\bottomrule
\end{tabular}
\caption{Comparison of existing AES approaches.}
\label{tab:aes_comparison}
\end{table*}

\subsection{From Traditional Feature-based to Deep Learning-based AES}
Traditional AES methods predominantly relied on handcrafted linguistic features to evaluate essay quality, focusing on surface-level metrics such as essay length, lexical diversity, syntactic complexity, and error rates \citep{yannakoudakis2011new}. For English L2 learners, early systems targeted learner corpora, such as the Cambridge Learner Corpus, to detect grammatical and lexical errors common in non-native writing \citep{yannakoudakis2011new}. Other work explored discourse features, like cohesion markers, to improve holistic scoring \citep{dong2016automatic}. Handcrafted features can perform well even with limited data \cite{wang2022use}, but they demand substantial expert effort. They may also miss higher-level qualities: especially for low-proficiency L2 writers, simple surface features (like spelling error rates) tend to dominate scoring \cite{de2025challenges}, limiting the model’s ability to capture subtle discourse or content aspects.

Deep neural models learn essay representations automatically, reducing reliance on manual features \citep{wang2022use}. Architectures such as LSTMs, CNNs, and RCNNs have been applied to essays to capture sequential and local context \citep{uto2020neural,mayfield2020should}. More recently, transformer-based models have been fine-tuned for AES such as in \citep{yang2020enhancing} combined regression. For English L2, these were applied to learner-specific tasks, such as scoring TOEFL essays by modeling long-distance dependencies in non-native structures \citep{taghipour2016neural}. 

Subsequent advances incorporated pre-trained transformers, with BERT-based systems fine-tuned for regression, enhancing cross-prompt generalization \citep{yang2020enhancing}. Multi-scale representations, combining token, sentence, and document levels, further improved accuracy on L2 corpora by addressing varying essay lengths and complexities \citep{wang2022use}. 

Attention mechanisms were key in highlighting rubric-relevant elements, like coherence in IELTS tasks \citep{dong2017attention}. Recent enhancements include joint learning with ranking losses to align scores more closely with human judgments \citep{yang2020enhancing}. Despite strong results, deep learning AES often requires large labeled datasets, which are scarce for L2-specific rubrics, and may overfit to domain-specific patterns \citep{li2024automated}. These models have been pivotal in L2 assessment, offering better feedback than traditional methods but still may not fully explain the rationale behind their predictions.

\begin{figure*}[!ht]
    \centering
    \includegraphics[width=6.25in]{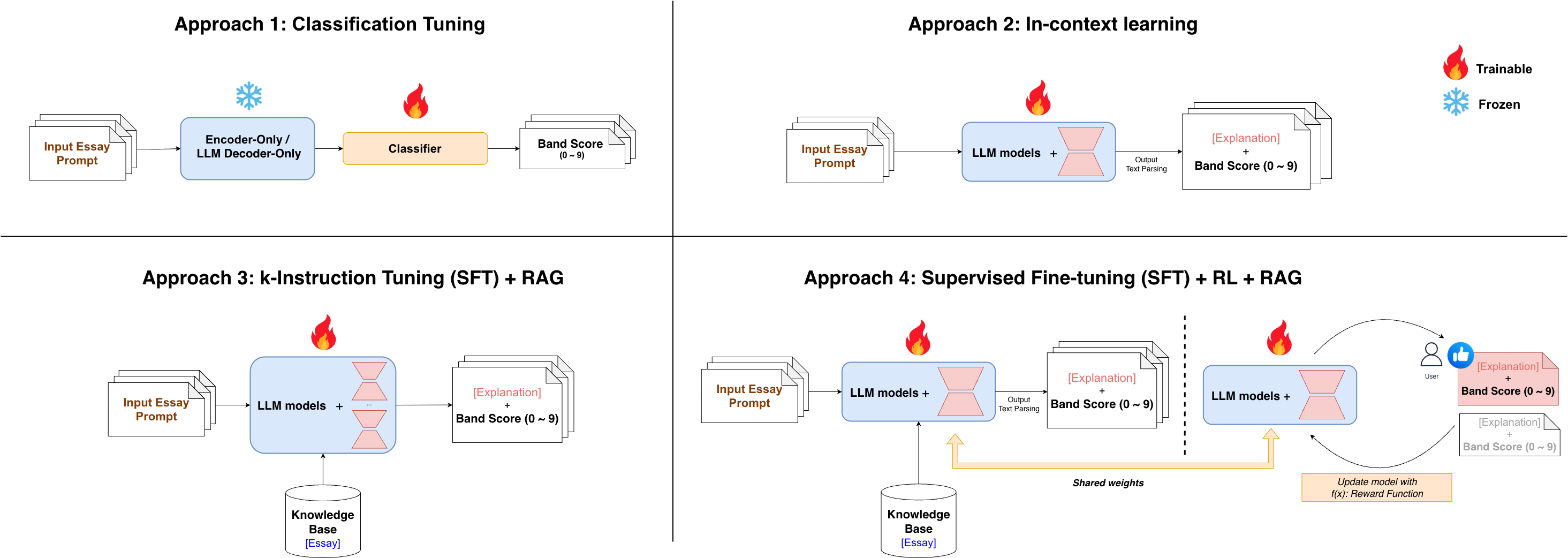}
    \caption{Overview of the four LLM-based AES paradigms evaluated in this study: (1) discriminative fine-tuning, (2) prompting-based inference, (3) instruction tuning with RAG, and (4) SFT with DPO and RAG.}
    \vspace{-4mm}
    \label{fig:overview}
\end{figure*}

\subsection{LLM-Based Approaches and IELTS Writing Assessment}
Large Language Models (LLMs) have recently been explored for essay scoring.  Prompting techniques, such as zero- or few-shot prompting, allow LLMs like GPT-4 to score essays without fine-tuning, achieving high correlations on IELTS and TOEFL benchmarks \citep{shibatamiyamura2025lces}. \cite{mizumoto2023exploring} used ChatGPT to grade TOEFL essays and found its scores moderately aligned with human raters. \cite{saricaoglu2025capacity} showed that ChatGPT-4 can accurately flag issues across key L2 writing dimensions (Task Response, Coherence, Vocabulary, Grammar) and provide relevant feedback. Other studies combine prompting and fine-tuning: \cite{qiu2024large} fine-tuned a RoBERTa encoder and separately used a GPT-4o prompt on IELTS Task 2 essays; both models achieved strong human–machine correlations (around 0.7) with official IELTS scores. Some recent work has also fine-tuned models for multi-dimensional IELTS scoring (e.g., \cite{sun2024automatic}), though such methods often rely on synthetic sub-scores. Preference optimization methods, like DPO, align models with human preferences, improving feedback quality for L2 learners \citep{li2024automated}. However, challenges remain in calibration and handling L2-specific errors, such as idiomatic usage \citep{song2025unified}. LLM-based AES promises scalable, explainable assessment for English L2, with recent work focusing on hybrid approaches combining prompting, fine-tuning, and augmentation for better generalization \citep{li2024automated, wang2025llms}.

AES research has evolved from feature-based to neural models to LLM-based methods (see Table~\ref{tab:aes_comparison}). These LLM-based approaches show promise for English L2 AES, but existing studies on L2 English AES are fragmented: most prior work targets either feature-based models or one form of LLM adaptation in isolation \cite{de2025challenges}. No comprehensive benchmark has compared fine-tuning, prompting, RAG, and preference-optimization strategies on the same L2 writing corpus. Our work fills this gap by systematically evaluating these methods on an IELTS writing dataset.

\section{Our methods}
Fig \ref{fig:overview} illustrates the overall framework of our LLM-based AES system, which systematically compares four representative adaptation paradigms for English L2 essay scoring. Given an IELTS Writing Task~2 prompt and a candidate essay, all approaches aim to predict the overall band score and, where applicable, analytic scores and feedback aligned with the official IELTS rubric. By evaluating all four paradigms under a unified experimental protocol and dataset, our framework enables a fair comparison of accuracy, and stability in LLM-based AES for English L2 writing.

\noindent\textbf{Approach 1: Discriminative fine-tunings.} Where encoder-based models (e.g., RoBERTa \cite{liu2019roberta}, GPT-2 \cite{radford2019language} encoder) are trained as classification models directly mapping essays to band scores. This approach serves as a strong neural AES baseline.

\noindent\textbf{Approach 2: In-context learning.} Leveraging large generative LLMs under zero-shot and few-shot prompting \cite{qiu2024large}. Essays are evaluated by explicitly instructing the model to act as an IELTS examiner, producing scores without task-specific parameter updates.

\noindent\textbf{Approach 3: k-Instruction tuning with RAG.} In this setting, the model is fine-tuned on 4-aspects instructions in IELTS Writing task 2, while external exemplars and scoring guidelines are retrieved to ground predictions and reduce hallucination.

\noindent\textbf{Approach 4: SFT with DPO and RAG.} This hybrid approach aligns model outputs with human scoring preferences and rubric consistency, aiming to improve robustness and calibration across proficiency levels.

\subsection{Problem Formulation}

We formulate Automated Essay Scoring (AES) for English L2 as a joint task of \emph{score prediction} and \emph{feedback generation}. 
Given an essay written by a second-language learner, the system aims to assign a proficiency score aligned with human raters while generating pedagogical feedback that explains the assigned score. Let $e_i$ denote an essay and $p_i$ its corresponding writing prompt.
The AES system produces a score--feedback pair:
\begin{equation}
    ( \hat{y}_i, f_i ) = \mathcal{M}_{\theta}(e_i, p_i, r_i),
\end{equation}
where $\hat{y}_i \in \mathcal{Y}$ is the predicted IELTS band score, $f_i \in \mathcal{F}$ is natural-language feedback, and $r_i$ denotes an optional rubric description.

\vspace{1mm}
\noindent\textbf{Discriminative Fine-Tuning.} AES is modeled as a supervised classification problem (Approach 1). Given an essay $E$ and its prompt $P$, the input sequence is constructed as:
\begin{equation}
    X_i = [\texttt{[CLS]}; P; \texttt{[SEP]}; E; \texttt{[SEP]}],
\end{equation}
where \texttt{[CLS]} is a classification token and \texttt{[SEP]} denotes segment boundaries.
An encoder produces a contextual representation $\mathbf{h}_{\texttt{[CLS]}}$, which is mapped to a discrete score prediction:
\begin{equation}
    \hat{y}_i = f_{\theta}(X_i), \quad \hat{y}_i \in \mathcal{Y}.
\end{equation}
The label space is defined as:
\begin{equation}
    \mathcal{Y} = \{0, 0.5, 1.0, \ldots, 8.5, 9.0\},
\end{equation}
resulting in a $K=19$-class classification setting.
This paradigm produces scores only and does not explicitly generate feedback.

\vspace{1mm}
\paragraph{LLM-Based Scoring and Feedback Generation.}
For generative LLM-based approaches (Approaches~2--4), Automated Essay Scoring is formulated as a conditional text generation problem.
Given an essay $e_i$, its corresponding prompt $p_i$, and optional rubric or retrieved context $r_i$, the model generates a structured output:
\begin{equation}
    o_i \sim p_{\theta}(o \mid e_i, p_i, r_i),
\end{equation}
where $p_{\theta}$ denotes a pretrained or fine-tuned generative language model, and $o_i$ is a textual output constrained by a predefined schema (e.g., JSON).

The generated output $o_i$ jointly encodes: (i) the predicted IELTS band score, and (ii) explanatory feedback aligned with official IELTS Writing Task~2 criteria. To enable quantitative evaluation, we define a deterministic parsing function $\phi(\cdot)$ that extracts the numerical score and feedback:
\begin{equation}
    (\hat{y}_i, f_i) = \phi(o_i),
\end{equation}
where $\hat{y}_i \in \mathcal{Y}$ is the predicted overall band score and $f_i$ denotes the associated textual feedback.

\subsection{Approach 1: Discriminative Fine-Tunings}
\label{sec:approach1}

\begin{figure}[htb!]
    \centering
    \includegraphics[width=\columnwidth]{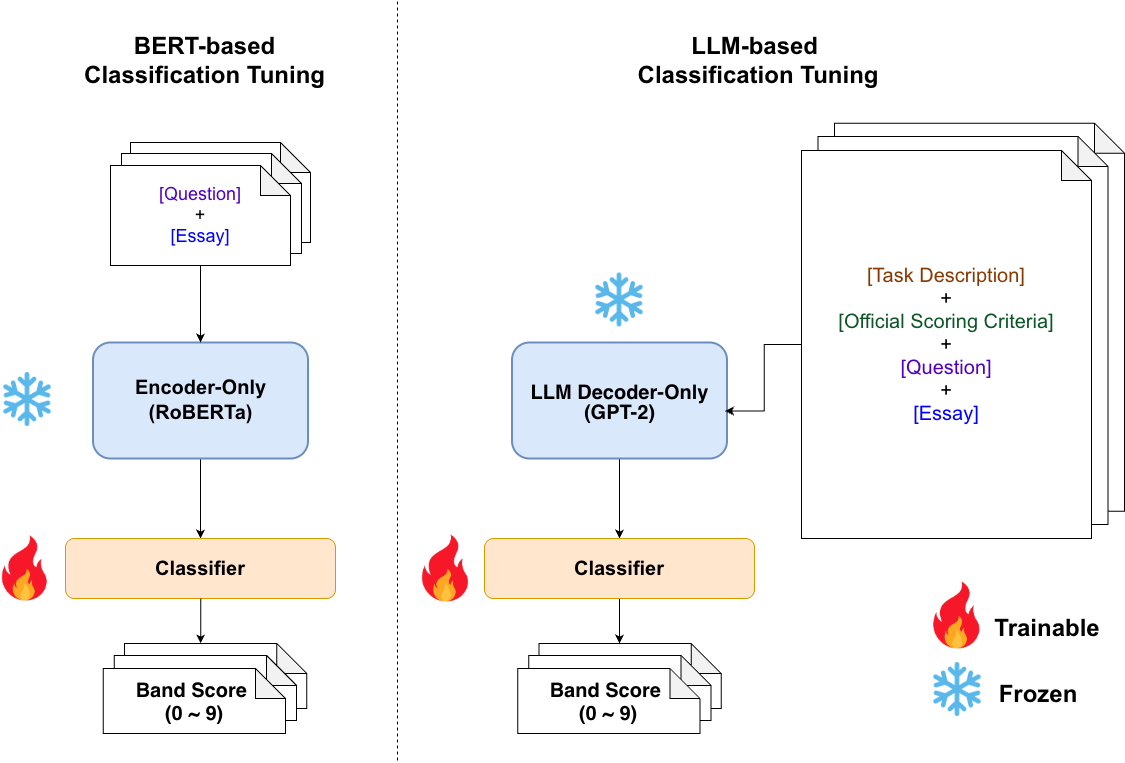}
    \caption{Overview of the discriminative fine-tuning approach for IELTS essay scoring.}
    \label{fig:approach_1}
\end{figure}

As a strong neural baseline, we formulate IELTS Writing Task~2 scoring as a supervised discriminative learning problem, where the model directly maps an essay (optionally concatenated with its prompt) to a discrete band score (Fig~\ref{fig:approach_1}).

\paragraph{Encoder-based Architecture.} We adopt pre-trained transformer encoders, including RoBERTa~\citep{liu2019roberta} (see \textbf{Study One}) and a GPT-2 encoder variant~\citep{radford2019language} (see \textbf{Study Two}), to obtain a contextualized representation of the input sequence:

\begin{equation}
    \mathbf{H} = \text{Encoder}(X)
\end{equation}

\noindent where $\mathbf{H}_{\texttt{[CLS]}} \in \mathbb{R}^{d}$ denotes the hidden state corresponding to the \texttt{[CLS]} token. With Study Two, we use Prompt in Appendix~\S\ref{subsec:criterion_prompt}.

On top of the encoder, we add a task-specific classification head composed of two fully connected layers:
\begin{align}
    \mathbf{z} &= \tanh(\mathbf{W}_1 \mathbf{H}_{\texttt{[CLS]}} + \mathbf{b}_1) \\
    \mathbf{o} &= \mathbf{W}_2 \mathbf{z} + \mathbf{b}_2
\end{align}
where $\mathbf{W}_1 \in \mathbb{R}^{h \times d}$, $\mathbf{W}_2 \in \mathbb{R}^{K \times h}$, and $h$ is the hidden dimension of the classification head. The final prediction probabilities are obtained via a softmax function:
\begin{equation}
    p(y = k \mid X) = \frac{\exp(o_k)}{\sum_{j=1}^{K} \exp(o_j)}
\end{equation}

\paragraph{Training Objective.}
We fine-tune only fully connected layers parameters for classification by minimizing the categorical cross-entropy loss \cite{bishop2006pattern}:

\begin{equation}
    \mathcal{L}_{\text{CE}} = - \sum_{k=1}^{K} \mathbb{I}(y = k) \log p(y = k \mid X)
\end{equation}
where $\mathbb{I}(\cdot)$ is the indicator function. This objective encourages the model to align its predictions with the discrete IELTS band labels.

\subsection{Approach 2: In-context Learning}
\label{sec:approach2}

\begin{figure*}[!htb!]
    \centering
    \includegraphics[width=6.25in]{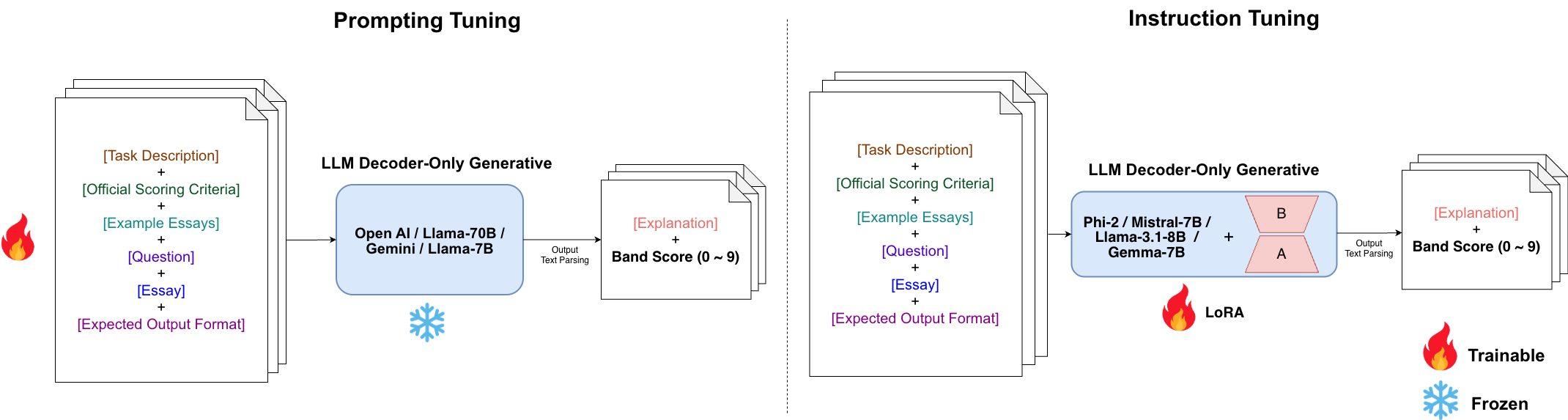}
    \caption{In-context learning paradigm for IELTS Writing Task~2 scoring, including prompting-based inference and instruction-tuned generative LLMs.}
    \label{fig:approach_2}
\end{figure*}

This approach investigates \emph{in-context learning} (ICL) as a generative alternative to discriminative fine-tuning for Automated Essay Scoring (Fig~\ref{fig:approach_2}).
In this paradigm, Large Language Models predict IELTS band scores by conditioning on natural-language instructions and optional examples, rather than relying on a task-specific classifier head.

We distinguish two conceptually different settings within this paradigm: 
\textbf{(i) prompting-based inference without parameter updates} and 
\textbf{(ii) instruction tuning with task-specific supervision}. 
Although both rely on natural-language instructions, they differ fundamentally in whether model parameters are optimized.

\subsubsection{Study Three: Prompting-based In-context Learning}
\label{subsec:study3_prompting}

In Study Three, we evaluate prompting-based ICL, where all model parameters $\theta$ remain \emph{frozen}. 
The model is instructed to act as an IELTS Writing Task~2 examiner and directly generate a band score, optionally conditioned on a small number of in-context examples. With this study, we use Prompt in Appendix~\S\ref{subsec:final_band_prompt}.

Given a prompt template $\mathcal{P}_{\text{ICL}}$ and a set of $k$ in-context exemplars $\mathcal{C} = \{(E_i, y_i)\}_{i=1}^k$, inference is performed as:
\begin{equation}
    o = f_\theta\bigl([\mathcal{P}_{\text{ICL}};\mathcal{C};P;E]\bigr),
\end{equation}
followed by:
\begin{equation}
    (\hat{y}, \hat{f}) = \phi(o),
\end{equation}
where $f_\theta$ denotes the autoregressive decoding function of the LLM.

We evaluate both zero-shot ($k=0$) and few-shot ($k \in \{2,4\}$) settings using closed-source (GPT-4o~\cite{hurst2024gpt}, Gemini~2.5~Pro~\cite{comanici2025gemini}) and open-source (Llama-3-70B~\cite{dubey2024llama}) models. This setting reflects the lowest adaptation cost scenario, requiring no training data and no gradient updates, but relying entirely on the model’s pretrained reasoning and instruction-following capabilities.

\begin{figure*}[htb!]
    \centering
    \includegraphics[width=6.25in]{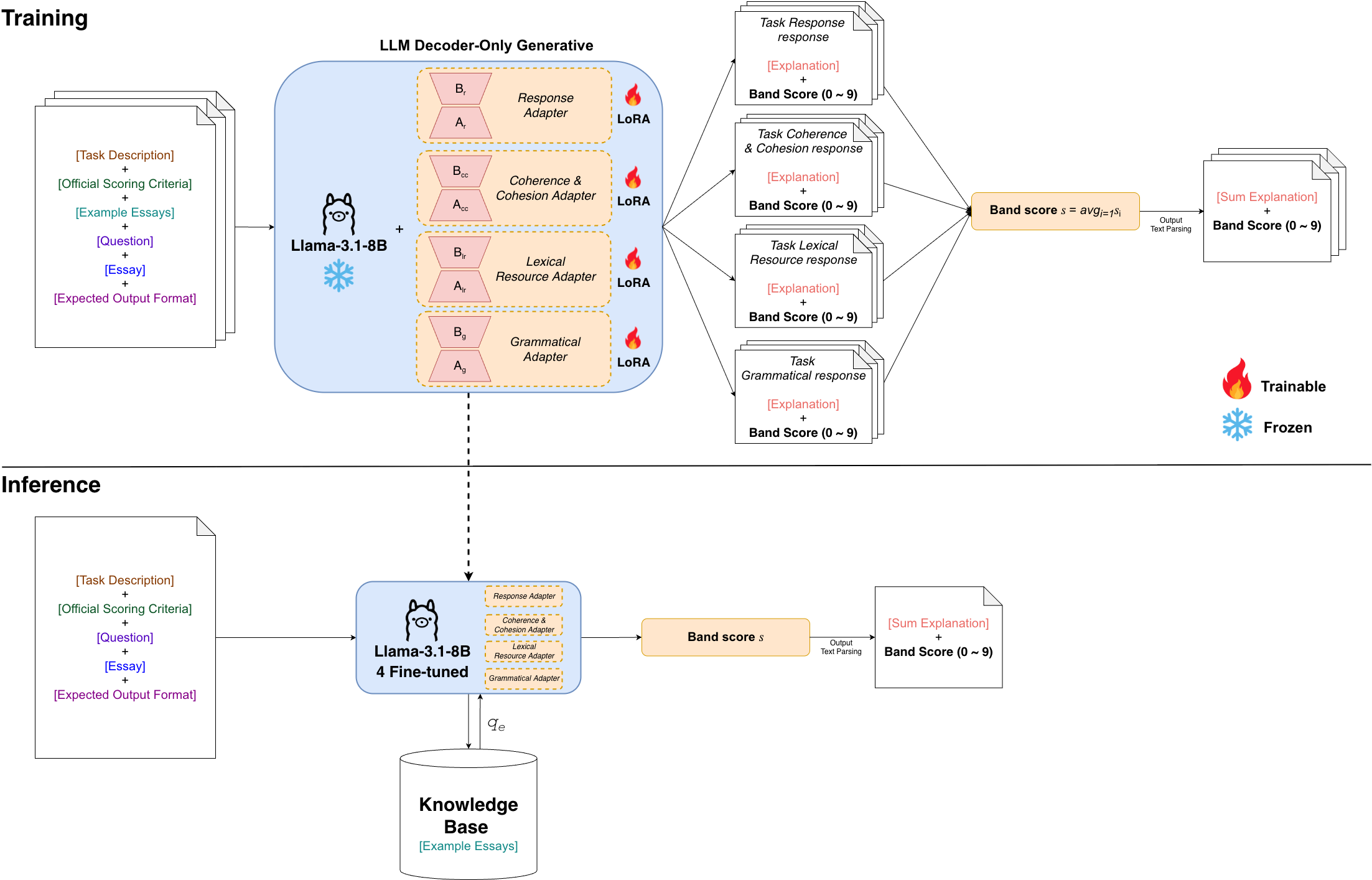}
    \caption{Overview of \textit{k}-instruction tuning with Retrieval-Augmented Generation (RAG) for IELTS Writing Task~2 scoring. The model is instruction-tuned on criterion-specific subtasks using LoRA adapters, while external rubric descriptions and exemplar essays are retrieved to ground inference and reduce hallucination.}
    \label{fig:approach_3}
\end{figure*}

\subsubsection{Study Four: Instruction Tuning}
\label{subsec:study4_instruction}

Study Four extends in-context learning by introducing \emph{instruction tuning}, where model parameters are explicitly optimized on task-specific supervision.
Given a training set of instruction–response pairs:
\begin{equation}
    \mathcal{D}_{\text{inst}} = \{(I_j, P_j, E_j, o_j)\}_{j=1}^{N},
\end{equation}
where $I_j$ denotes a criterion-specific instruction and $o_j$ is a structured target output containing both the predicted score and explanatory feedback, the model is trained to maximize the conditional likelihood of $o_j$. With this study, we use Prompt in Appendix~\S\ref{subsec:final_band_prompt}.

To improve parameter efficiency, we adopt Low-Rank Adaptation (LoRA)~\cite{hu2022lora}.
Specifically, for each adapted weight matrix $W \in \mathbb{R}^{d \times k}$ in the backbone LLM, LoRA reparameterizes it as:
\begin{equation}
    W' = W + \Delta W, \quad \Delta W = B A,
\end{equation}
where $A \in \mathbb{R}^{r \times k}$ and $B \in \mathbb{R}^{d \times r}$ are low-rank trainable matrices with rank $r \ll \min(d,k)$, while the original weight $W$ remains frozen.

Let $\theta$ denote the frozen backbone parameters and $\phi$ the set of LoRA parameters.
Instruction tuning then minimizes the negative log-likelihood with respect to $\phi$:
\begin{equation}
    \mathcal{L}_{\text{IT}} = - \sum_{j=1}^{N} 
    \log p_{\theta,\phi}(o_j \mid I_j, P_j, E_j),
\end{equation}
where $p_{\theta,\phi}$ denotes the generative model with LoRA-adapted weights.

We evaluate instruction tuning under zero-shot ($k=0$) and few-shot ($k \in \{2,4\}$) prompting settings using a diverse set of open-source instruction-following LLMs, including Gemma-7B~\citep{team2024gemma}, Phi-2~\citep{javaheripi2023phi}, Mistral-7B~\citep{jiang2023mistral}, and Llama-3.1-8B~\citep{dubey2024llama}. 
All models are trained or evaluated using identical instruction templates and constrained decoding schemes to generate structured outputs containing both band scores and feedback.
This setting captures a moderate adaptation-cost regime, where task-specific knowledge is incorporated via supervised instruction tuning while retaining the flexibility of in-context calibration at inference time.

\subsection{Approach 3: \textit{k}-Instruction Tuning with Retrieval-Augmented Generation}
\label{sec:approach3}

This approach extends instruction tuning (Approach~2) by integrating \emph{Retrieval-Augmented Generation (RAG)} and \emph{criterion-specific parameter-efficient fine-tuning} (Fig~\ref{fig:approach_3}). 
The key idea is to decompose IELTS Writing Task~2 assessment into $k=4$ official evaluation criteria—\textit{Task Response (TR)}, \textit{Coherence and Cohesion (CC)}, \textit{Lexical Resource (LR)}, and \textit{Grammatical Range and Accuracy (GRA)}—and to explicitly ground inference using retrieved aligned essay. 

Let $E_i$ denote an essay and $P_i$ its corresponding prompt, the model input is augmented with retrieved context:
\begin{equation}
    o_i \sim p_{\theta}(o \mid I_i, P_i, E_i, \mathcal{R}(E_i)),
\end{equation}
where $I_i$ is a criterion-specific instruction and $\mathcal{R}(E_i)$ retrieves relevant descriptions or exemplar essays from an external corpus.
The generated output $o_i$ is a structured response containing both a criterion-level score and explanatory feedback. This retrieval grounding improves adherence, mitigates hallucination, and stabilizes score calibration, particularly for borderline proficiency levels.

\begin{figure*}[htb!]
    \centering
    \includegraphics[width=6.25in]{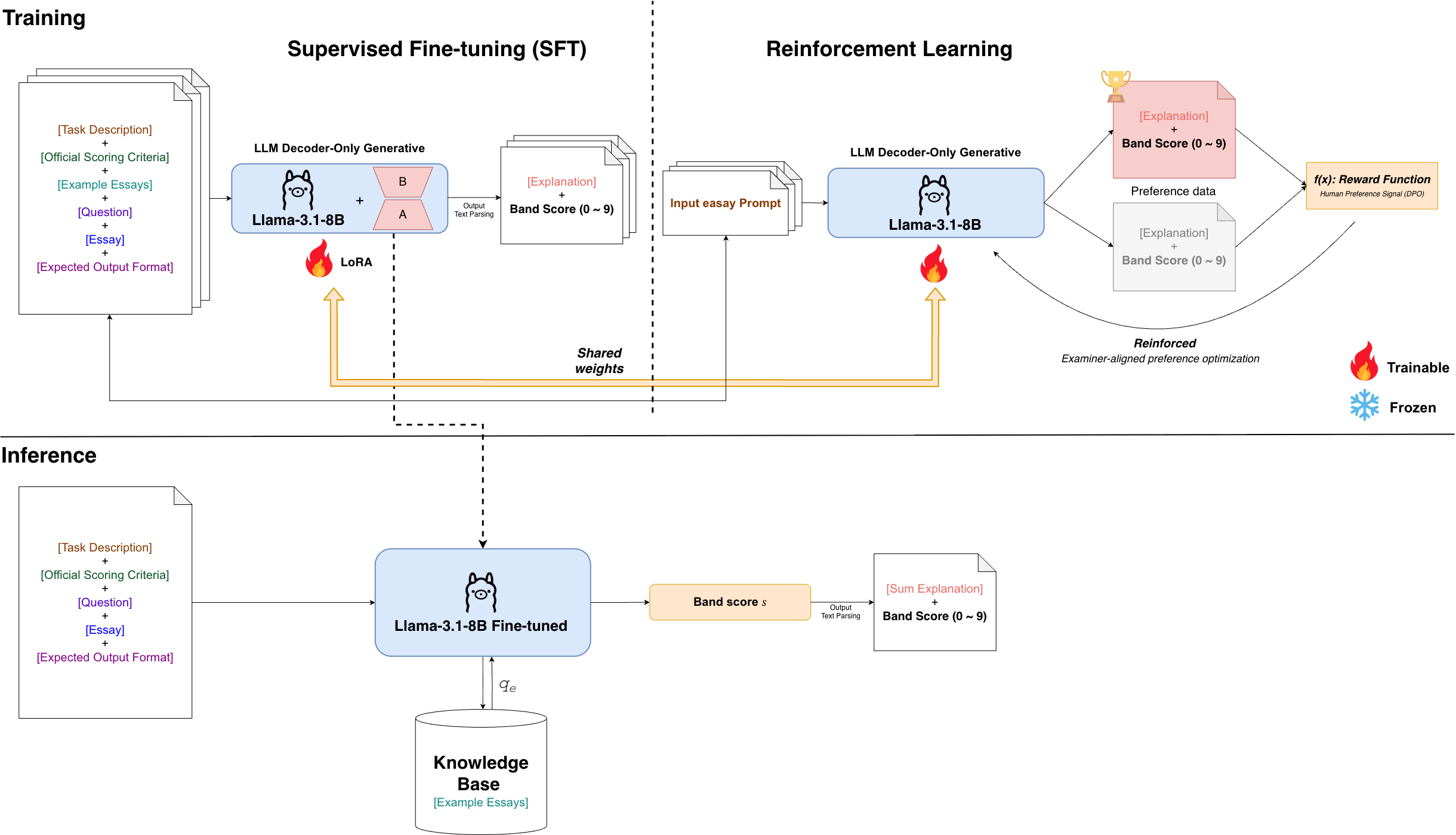}
    \caption{Overview of Supervised Fine-Tuning with Direct Preference Optimization (DPO) and Retrieval-Augmented Generation (RAG). The model is first fine-tuned to generate score--feedback pairs, then aligned using human preference data, and finally grounded with retrieved exemplars at inference time.}
    \label{fig:approach_4}
\end{figure*}

\paragraph{\textit{k}-Instruction Tuning via Criterion-Specific LoRA.}
To enable targeted specialization while maintaining parameter efficiency, we adopt Low-Rank Adaptation (LoRA)~\cite{hu2022lora} and fine-tune \textbf{four independent LoRA adapters}, one for each IELTS criterion.
Each adapter is trained on the same backbone model (Llama-3.1-8B~\citep{dubey2024llama}) but receives a distinct instruction prompt aligned with a single evaluation aspect.

Formally, for criterion $c \in \{\text{TR}, \text{CC}, \text{LR}, \text{GRA}\}$, the LoRA-adapted model generates:
\begin{equation}
    o_i^{(c)} \sim p_{\theta + \Delta\theta^{(c)}}\!\left(o \mid I_i^{(c)}, P_i, E_i, \mathcal{R}(E_i)\right),
\end{equation}
where $\Delta\theta^{(c)}$ denotes the low-rank parameter update for criterion $c$.
Each output $o_i^{(c)}$ is constrained to a strict JSON format containing a band score and criterion-specific feedback.

The final overall band score is computed by aggregating the four criterion scores:
\begin{equation}
    \hat{y}_i = \frac{1}{4} \sum_{c} \phi_{\text{score}}\!\left(o_i^{(c)}\right),
\end{equation}
where $\phi_{\text{score}}(\cdot)$ extracts the numerical band score from each criterion-level output.


We evaluate both non-fine-tuned and instruction-tuned models under a unified RAG framework~\citep{lewis2020retrieval}, including Mistral-7B-Instruct-v0.3~\citep{jiang2023mistral} and Llama-3.1-8B~\citep{dubey2024llama}. Depending on the configuration, models are prompted either to perform direct final-band prediction or criterion-level scoring, as specified in Appendix~\S\ref{subsec:final_band_prompt} and Appendix~\S\ref{subsec:criterion_prompt}. All RAG-based experiments use a two-shot ($k=2$) retrieval setting. Detailed prompt templates for each criterion are provided in Appendix~\S\ref{subsec:single_criterion}.

\subsection{Approach 4: Supervised Fine-Tuning with Reinforcement Learning and RAG}

This approach combining \emph{supervised fine-tuning (SFT)}, \emph{preference-based reinforcement learning}, and \emph{retrieval-augmented generation (RAG)} (Fig~\ref{fig:approach_4}).
The goal is to jointly improve score accuracy, feedback quality, and faithfulness by aligning the model with both labeled supervision and human judgment.

\paragraph{Supervised Fine-Tuning.}
We first perform supervised fine-tuning using instruction--response pairs, where the model is trained to directly generate structured outputs containing both an overall band score and detailed feedback. Given an essay $E_i$ and prompt $P_i$, the model generates:
\begin{equation}
    (\hat{y}_i, f_i) \sim p_{\theta}(y, f \mid P_i, E_i),
\end{equation}
where $\hat{y}_i$ denotes the predicted IELTS band score and $f_i$ represents explanatory feedback.
The SFT stage uses the \emph{Criterion-level Prompt} described in Appendix~\S\ref{subsec:criterion_prompt}.

\paragraph{Preference Alignment via Direct Preference Optimization.}
While SFT improves factual correctness, it does not explicitly optimize for feedback quality or human preference.
To address this limitation, we further refine the model using \emph{Direct Preference Optimization (DPO)}, a reinforcement learning method that aligns model outputs with pairwise human preferences.

Given a triplet $(E_i, f_i^{+}, f_i^{-})$, where $f_i^{+}$ is a preferred feedback and $f_i^{-}$ is a dispreferred one, DPO optimizes the model to increase the likelihood of preferred responses:
\begin{equation}
    p_{\theta}(f_i^{+} \succ f_i^{-} \mid P_i, E_i).
\end{equation}
We leverage the publicly available \texttt{IELTS\_essay\_human\_feedback} dataset\footnote{\url{https://huggingface.co/datasets/chillies/IELTS_essay_human_feedback}}, which provides human-annotated preference pairs over feedback quality.
This stage encourages clearer explanations, better error localization, and more constructive suggestions, beyond what is achievable with likelihood-based training alone.

\paragraph{Retrieval-Augmented Inference.}
At inference time, we integrate Retrieval-Augmented Generation to ground predictions in external evidence.
For a given essay, relevant exemplar essays and rubric descriptions are retrieved and appended to the model input:
\begin{equation}
    (\hat{y}_i, f_i) \sim p_{\theta}(y, f \mid P_i, E_i, \mathcal{R}(E_i)).
\end{equation}
This grounding mechanism reduces hallucination, improves consistency across proficiency levels, and stabilizes final band predictions, particularly for borderline cases.

We evaluate two instruction-following LLMs under the same optimization pipeline: Mistral-7B-Instruct-v0.3~\citep{jiang2023mistral} and Llama-3.1-8B~\citep{dubey2024llama}.
For both models, we apply a single parameter-efficient LoRA~\cite{hu2022lora} adapter during the supervised fine-tuning (SFT) and Direct Preference Optimization (DPO)~\cite{rafailov2023direct} stages, allowing efficient adaptation while preserving the pretrained backbone.
At inference time, Retrieval-Augmented Generation is employed with a two-shot ($k=2$) retrieval setting, consistent with the configurations used in Approaches~2 and~3.

Together, Approaches~2–4 illustrate a principled progression from parameter-free prompting to fully aligned, retrieval-grounded optimization, highlighting the complementary roles of instruction tuning, preference learning, and external evidence in LLM-based Automated Essay Scoring.

\section{Experiments}

\subsection{Dataset Construction}

\noindent\textbf{Data Sources.}
Our experiments are conducted on a consolidated IELTS Writing Task~2 dataset constructed from two publicly available sources, which play complementary roles in our data pipeline. The primary source is the \textit{IELTS Writing Task~2 Evaluation Dataset} released on Hugging Face by chillies%
\footnote{\url{https://huggingface.co/datasets/chillies/IELTS-writing-task-2-evaluation}}, containing 10,324 essay samples. Each instance includes a writing prompt, an essay response, and an overall band score, together with automatically generated analytic evaluations across four official IELTS criteria: Task Response (TR), Coherence and Cohesion (CC), Lexical Resource (LR), and Grammatical Range and Accuracy (GRA). The second source is the \textit{Raw IELTS Essays} dataset released on Kaggle by arsenycheplukov%
\footnote{\url{https://www.kaggle.com/datasets/arsenycheplukov/raw-ielts-essays}}, comprising 6,944 essays with overall band scores, CEFR levels, and human-written feedback. This dataset is \emph{not directly merged} into the final corpus. Instead, it is used as an auxiliary resource for prompt calibration, rubric grounding, and qualitative validation of analytic criteria during prompt design.

\vspace{1mm}
\noindent\textbf{Re-generation of Analytic Scores.}
During preliminary inspection, we observed that the analytic evaluation fields in the Hugging Face dataset were produced using an earlier Gemini model under loosely constrained prompts, resulting in occasional inconsistencies between analytic criteria and the provided overall band scores. To improve internal consistency and rubric alignment, we re-generated analytic scores using the \texttt{Gemini-2.5-Pro} API model~\cite{comanici2025gemini} with a strictly controlled prompting protocol. The prompt design enforces: (i) explicit adherence to official IELTS Writing Task~2 band descriptors; (ii) conditioning on the known overall band score as a soft constraint; and (iii) structured JSON outputs for reliable downstream processing. Full prompt details are provided in Appendix~\S\ref{sec:appendix_regen_prompt}.

\begin{figure*}[ht]
\centering
\begin{subfigure}[t]{0.49\linewidth}
    \centering
    \includegraphics[width=\linewidth]{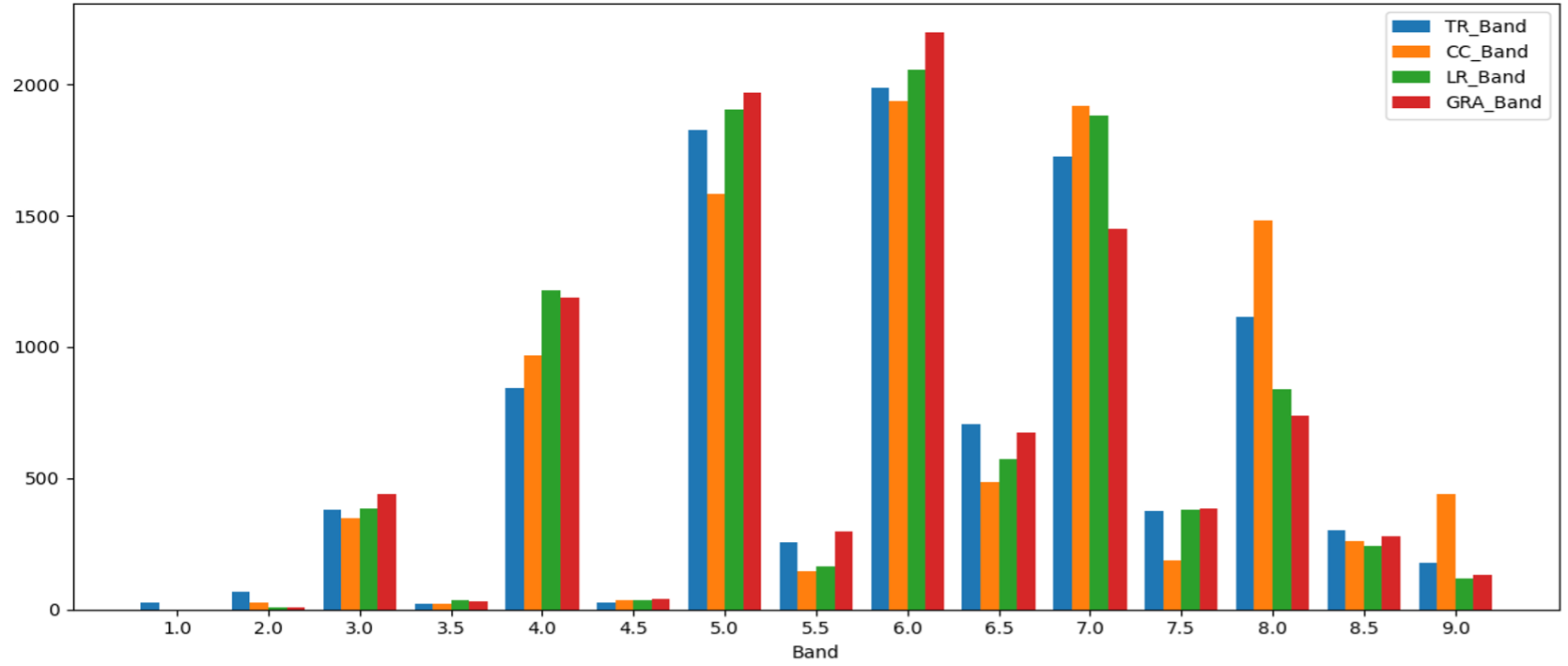}
\end{subfigure}
\hfill
\begin{subfigure}[t]{0.46\linewidth}
    \centering
    \includegraphics[width=\linewidth]{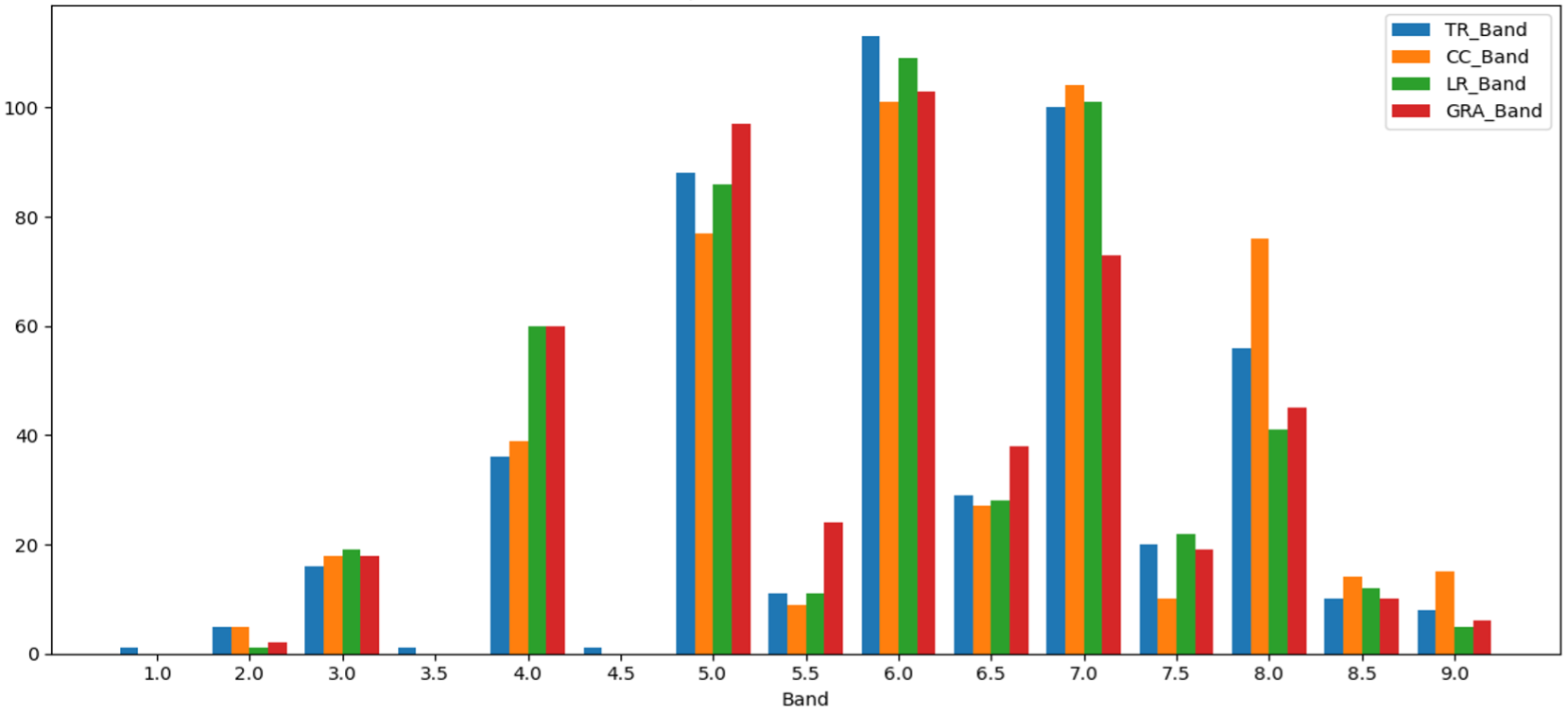}
\end{subfigure}
\caption{Distribution of analytic criterion scores in the dataset: (left) training set, (right) test set.}
\label{fig:score_dist_analytic}
\end{figure*}

\vspace{1mm}
\noindent\textbf{Dataset Analysis.} After filtering incomplete entries, resolving formatting issues, and consolidating regenerated analytic fields, the final dataset contains 10,328 essays\footnote{\textit{Most of analytic scores are automatically generated by a large language model based on IELTS band descriptors rather than assigned by certified human examiners. These labels are used as weakly supervised reference signals for large-scale analysis, not as gold-standard human judgments.}}. We adopt a fixed split of 9,833 essays for training and 495 essays for testing. The test set is held out across all experiments to ensure fair and reproducible comparison among different AES paradigms.

\begin{table}[ht]
    \centering
    \small
    \setlength{\tabcolsep}{6pt}
    \begin{tabular}{lcc}
    \hline
    \textbf{Statistic} & \textbf{Train} & \textbf{Test} \\
    \hline
    \# Essays            & 9,833 & 495 \\
    Avg. Essay Length (tokens) & 231 & 228 \\
    Score Range         & 0--9 & 0--9 \\
    Mean Overall Score  & 5.5 & 5.5 \\
    Std. Dev.           & 1.2 & 1.1 \\
    \hline
    \end{tabular}
    \caption{Dataset statistics for training and test splits.}
    \label{tab:dataset_statistics}
\end{table}

\begin{figure}[ht]
\centering
\begin{subfigure}[t]{0.48\linewidth}
    \centering
    \includegraphics[width=\linewidth]{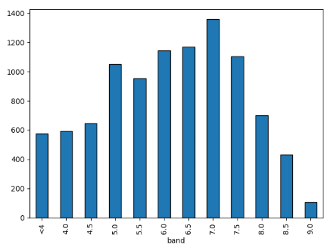}
\end{subfigure}
\hfill
\begin{subfigure}[t]{0.48\linewidth}
    \centering
    \includegraphics[width=\linewidth]{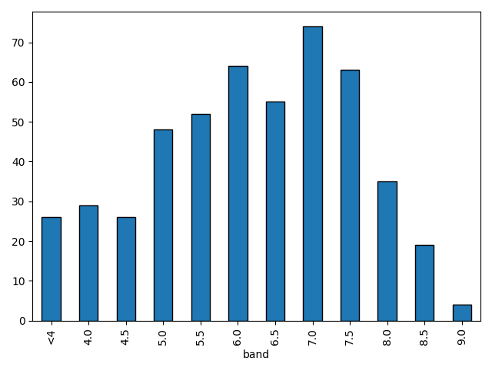}
\end{subfigure}
\caption{Distribution of overall band scores in the dataset: (left) training set, (right) test set.}
\label{fig:score_dist_overall}
\end{figure}

Fig \ref{fig:score_dist_overall} shows the distribution of overall band scores for the training and test sets. The two distributions exhibit highly similar shapes and ranges, indicating that the data partitioning does not introduce noticeable score imbalance across levels.

Fig \ref{fig:score_dist_analytic} presents the distributions of the four analytic criteria (TR, CC, LR, GRA) for the training and test sets. The distributions largely overlap between splits, suggesting comparable coverage of proficiency levels for each criterion. As with the overall scores, these analytic labels are automatically produced by an LLM and should be interpreted as approximate indicators of scoring tendencies rather than precise human judgments.

\subsection{Experimental Settings}
\label{subsec:exp_setting}

All experiments are conducted under a unified training and evaluation protocol to ensure comparability across adaptation strategies. Discriminative fine-tuning (Approach~1: Studies~1--2), in-context learning (Approach~2: Studies~3--4), \textit{k}-instruction tuning with Retrieval-Augmented Generation (Approach~3: Study~5), and supervised fine-tuning with preference optimization (Approach~4: Study~6) are evaluated on the same dataset splits and evaluation metrics. Unless otherwise specified, optimization is performed using AdamW~\cite{loshchilovdecoupled} with linear learning-rate scheduling. Parameter-efficient training is implemented via Low-Rank Adaptation (LoRA) with rank $r{=}16$, trained for three epochs with a per-device batch size of~2 and gradient accumulation. All experiments are run on a single NVIDIA Tesla T4 GPU, employing memory-efficient techniques such as 4-bit quantization, gradient checkpointing, and 8-bit optimizers where applicable. Detailed hyperparameters and training configurations for each study are provided in Appendix~\ref{sec:appendix_hyperparams}.

\subsection{Main Results}
\label{subsec:main_results}

\begin{table*}[!ht]
\centering
\small
\setlength{\tabcolsep}{5pt}
\begin{tabular}{
p{1.3cm}
p{4cm}
p{2.8cm}
c
c c c c
}
\toprule
\textbf{Approach} &
\textbf{Model / Setting} &
\textbf{Training / Prompting Scheme} &
\textbf{k-shot} &
\textbf{Accuracy} &
\textbf{F1} &
\textbf{RMSE} &
\textbf{MAE} \\
\midrule

\multirow{4}{*}{\textbf{A1}}
& RoBERTa-base (BERT-based) & Classifier Only        & --        & 0.6510 & 0.2160 & 0.8300 & 0.5810 \\
& RoBERTa-base (BERT-based) & All Parameters         & --        & 0.7310 & 0.3040 & 0.7840 & 0.5490 \\
& GPT-2 (LLM-based)         & Classifier Only        & --        & 0.7070 & 0.2780 & 0.7570 & 0.5300 \\
& GPT-2 (LLM-based)         & All Parameters         & --        & 0.7350 & 0.3090 & 0.7700 & 0.5390 \\
\midrule

\multirow{5}{*}{\textbf{A2}}
& GPT-4o                    & Prompting Tuning       & 2-shot    & 0.7100 & 0.2810 & 1.0500 & 0.7350 \\
& GPT-4o                    & Prompting Tuning       & zero-shot & 0.7200 & 0.2920 & 1.1300 & 0.7910 \\
& Llama-3-70B               & Prompting Tuning       & 2-shot    & 0.5600 & 0.1160 & 1.2500 & 0.8750 \\
& Llama-3-70B               & Prompting Tuning       & zero-shot & 0.6300 & 0.1930 & 0.9900 & 0.6930 \\
& Gemini 2.5 Pro            & Prompting Tuning       & zero-shot & 0.9616 & 0.5581 & 1.4977 & 1.2026 \\
\midrule

\multirow{6}{*}{\textbf{A3}}
& Gemma-7B                  & Instruction Tuning            & zero-shot & 0.6700 & 0.2200 & 1.3300 & 0.9800 \\
& Phi-2                     & Instruction Tuning            & zero-shot & 0.7812 & 0.4100 & 1.3500 & 0.6505 \\
& Mistral-7B                & Instruction Tuning            & zero-shot & 0.7000 & 0.2700 & 1.4128 & 1.1214 \\
& Llama-3.1-8B              & Instruction Tuning            & zero-shot & 0.7596 & 0.4511 & 1.0172 & 0.9859 \\
& Llama-3.1-8B              & Instruction Tuning + Few-shot & 2-shot    & 0.8680 & 0.5200 & 0.9400 & 1.0200 \\
& Llama-3.1-8B              & Instruction Tuning + Few-shot & 4-shot    & 0.8708 & 0.5679 & 1.3547 & 0.9436 \\
\midrule

\multirow{5}{*}{\textbf{A3 + RAG}}
& Mistral-7B-Instruct-v0.3 (not finetuned) & k-Instruction Tuning + RAG & 2-shot & 0.9050 & 0.7200 & 1.1800 & 0.8500 \\
& Llama-3.1-8B (not finetuned)              & k-Instruction Tuning + RAG & 2-shot & 0.9630 & 0.8400 & 0.9800 & 0.7000 \\
& Llama-3.1-8B (1-LoRA)                     & k-Instruction Tuning + RAG & 2-shot & 0.9750 & 0.8800 & 0.9200 & 0.6600 \\
& Llama-3.1-8B (4-LoRA)                     & k-Instruction Tuning (w/o RAG) & 2-shot & 0.9818 & 0.7999 & 0.9947 & 0.7676 \\
& \textbf{Llama-3.1-8B (4-LoRA)}             & \textbf{k-Instruction Tuning + RAG} & \textbf{2-shot}
& \textbf{0.9902} & \textbf{0.9350} & \underline{0.8700} & \underline{0.6200} \\
\midrule

\multirow{2}{*}{\textbf{A4}}
& Mistral-7B-Instruct-v0.3 (1-LoRA) & SFT + DPO + RAG & 2-shot & 0.9470 & 0.8750 & 1.0300 & 0.8300 \\
& Llama-3.1-8B (1-LoRA)             & SFT + DPO + RAG & 2-shot
& \underline{0.9870} & \underline{0.9250} & \textbf{0.8400} & \textbf{0.5800} \\
\bottomrule
\end{tabular}

\caption{Main results on IELTS Writing Task~2 Automated Essay Scoring.
A1: Discriminative fine-tuning; 
A2: Prompting-based; 
A3: Instruction tuning; 
A4: Supervised fine-tuning with preference optimization.
Higher is better for Accuracy and F1-score, lower is better for RMSE and MAE. Bold best results, underline second-best.}
\label{tab:main_results_aes}
\end{table*}

Table~\ref{tab:main_results_aes} presents the consolidated results across all four AES paradigms evaluated in this study, including discriminative fine-tuning, prompting-based inference, instruction tuning, and supervised fine-tuning with preference optimization.

Overall, traditional discriminative baselines (Approach~1) yield moderate performance, with accuracy below 0.75 and relatively low F1-scores, reflecting limited robustness in capturing nuanced rubric-level distinctions. Prompting-based LLMs (Approach~2) improve flexibility and, in some cases, accuracy, but exhibit substantial variance across models and prompting strategies. Instruction-tuned and retrieval-augmented methods (Approach~3) substantially outperform prior paradigms, achieving strong gains across all metrics. The best-performing configuration—Llama-3.1-8B with k-instruction tuning and RAG—achieves an Accuracy of 0.9902 and an F1-score of 0.9350. Finally, supervised fine-tuning with preference optimization (Approach~4) further improves error-based metrics, yielding the lowest RMSE and MAE, indicating more calibrated and human-aligned predictions.

Across all experiments, we observe a clear performance hierarchy:
\emph{discriminative fine-tuning} $\;<\;$ \emph{prompting-based inference} $\;<\;$ \emph{instruction tuning} $\;<\;$ \emph{SFT with RAG and preference optimization} $\;<\;$ \emph{k-SFT with RAG}.
While prompting offers a low-cost entry point, instruction-tuned and retrieval-augmented models consistently deliver superior accuracy and stability, highlighting the importance of task-specific supervision and rubric grounding for reliable L2 essay assessment.

\paragraph{Effect of Prompting.}
Comparing encoder-based fine-tuning with prompting-based LLMs, we find that discriminative models provide relatively stable but limited performance, with F1-scores below 0.31.
Prompting-based methods improve flexibility and, in some cases, accuracy; however, their performance varies substantially across models and prompting strategies.
Closed-source models (e.g., GPT-4o, Gemini) generally outperform open-source counterparts under zero-shot settings but incur significantly higher inference costs and exhibit larger error variance (RMSE/MAE).
This highlights a trade-off between deployment cost and performance consistency.

\paragraph{Effect of Instruction Tuning with Task Specific.}
Instruction tuning (Study~4) consistently improves both accuracy and F1-score over pure prompting (Study~3), particularly for open-source LLMs.
For example, Llama-3.1-8B benefits substantially from few-shot instruction tuning, yielding more stable predictions and higher F1-scores. These results indicate that explicit task supervision enables LLMs to better internalize IELTS-specific scoring criteria, reducing sensitivity to prompt phrasing and example selection.

\begin{figure}[htb!]
    \centering
    \includegraphics[width=\columnwidth]{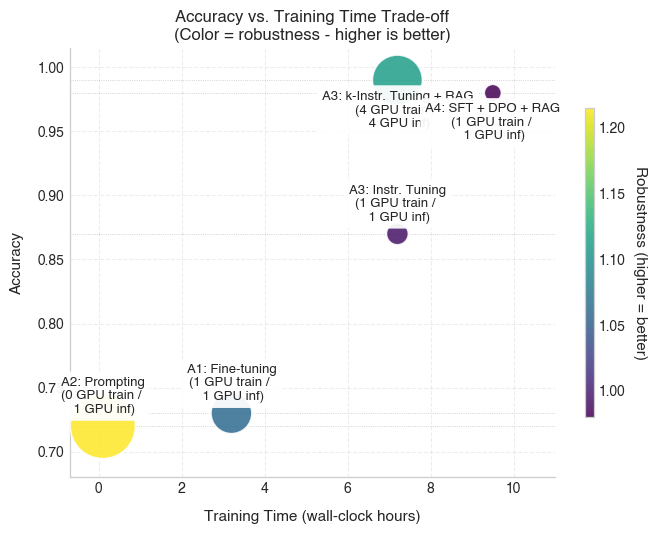}
    \caption{Cost--accuracy--robustness trade-offs across LLM-based AES paradigms for IELTS Writing Task~2.}
    \label{fig:cost_acc_tradeoff}
\end{figure}

\subsection{Cost--Accuracy--Robustness Trade-offs}
\label{subsec:cost_accuracy_tradeoff}

Figure~\ref{fig:cost_acc_tradeoff} visualizes the trade-offs between scoring accuracy and computational cost across five AES paradigms, where accuracy denotes overall classification performance and cost is approximated by wall-clock training time and GPU usage (see Training time and GPU usage in Appendix~\ref{sec:appendix_hyperparams}). Lightweight approaches such as discriminative fine-tuning (A1) and prompting (A2) incur minimal training cost (3.2 and 0.1 hours, respectively) but achieve limited accuracy (0.73 and 0.72), indicating poor scalability to high-quality assessment. Instruction tuning (A3) substantially improves accuracy to 0.87 at a moderate cost of 7.2 training hours on a single GPU, representing a favorable efficiency gain. The highest accuracy is obtained by \textit{k}-instruction tuning with RAG (A3+RAG), reaching 0.99 accuracy with comparable training time but higher GPU usage, highlighting the effectiveness of retrieval grounding despite increased resource demands. Finally, SFT with DPO and RAG (A4) achieves near-optimal accuracy (0.98) at the highest cost (9.5 hours), suggesting diminishing returns in accuracy but improved calibration. Overall, when ranking approaches by accuracy–cost efficiency, A3+RAG emerges as the most effective, followed by A3, A4, and the low-cost but less reliable A1 and A2.

\subsection{Case Study: k-Instruction Tuning vs. Reference Optimization}
\label{subsec:case_study}

\begin{table*}[!ht]
\centering
\small
\setlength{\tabcolsep}{6pt}
\begin{tabular}{p{1.7cm} p{2.5cm} p{1.1cm} p{2cm} p{6.4cm}}
\hline
\textbf{Topic} & \textbf{Essay Excerpt} & \textbf{Gold} & \textbf{Predicted Band} & \textbf{LLM Feedback (excerpt)} \\
\hline
Technology \& Education
& ``Some people think computers should replace teachers...'' 
& 6.5
& 
\textbf{A3+RAG}: \textcolor[HTML]{029e73}{6.5} \newline
\textbf{A4}: \textcolor[HTML]{029e73}{6.5}
&
\textbf{A3+RAG}: \newline
\emph{Task Response}: ...\textcolor[HTML]{029e73}{Addresses both views clearly and remains on-topic}, though \textcolor[HTML]{de8f05}{the evaluation in the conclusion is limited}... \newline
\emph{Coherence}: \textcolor[HTML]{029e73}{Paragraphing follows a logical structure}, with \textcolor[HTML]{de8f05}{minor lapses in cohesion}. \newline
\emph{Lexical}: \textcolor[HTML]{029e73}{Uses an adequate range of topic-related vocabulary}... \newline
\emph{Grammar}: \textcolor[HTML]{029e73}{Mostly accurate sentence forms}, despite \textcolor[HTML]{de8f05}{occasional tense errors}...\newline
\textbf{A4}: \newline
\textcolor[HTML]{029e73}{A clear and consistent position is maintained throughout the essay}. 
\textcolor[HTML]{029e73}{Ideas are generally well connected and easy to follow}, although 
\textcolor[HTML]{de8f05}{smoother transitions would strengthen coherence}...\\
\hline

Urbanization
& ``Living in big cities brings more benefits than problems...'' 
& 7.0
&
\textbf{A3+RAG}: \textcolor[HTML]{029e73}{7.0} \newline
\textbf{A4}: \textcolor[HTML]{de8f05}{6.5}
&
\textbf{A3+RAG}: \newline
\emph{Task Response}: \textcolor[HTML]{029e73}{Presents well-developed and balanced arguments}... \newline
\emph{Coherence}: \textcolor[HTML]{029e73}{Ideas progress clearly across paragraphs}. \newline
\emph{Lexical}: \textcolor[HTML]{029e73}{Demonstrates a wide range of topic-specific vocabulary}... \newline
\emph{Grammar}: \textcolor[HTML]{029e73}{Uses complex sentence structures with good control}... \newline
\textbf{A4}: \newline
...\textcolor[HTML]{de8f05}{While the main ideas are relevant, their development is uneven}; 
some paragraphs lack focus, which \textcolor[HTML]{de8f05}{results in a slightly lower overall band}... \\
\hline

Environmental Protection
& ``Governments should invest more in renewable energy...'' 
& 6.0
&
\textbf{A3+RAG}: \textcolor[HTML]{029e73}{6.0} \newline
\textbf{A4}: \textcolor[HTML]{029e73}{6.0}
&
\textbf{A3+RAG}: \newline
\emph{Task Response}: ...\textcolor[HTML]{de8f05}{Only partially addresses the task}, with limited supporting examples. \newline
\emph{Coherence}: ...\textcolor[HTML]{de8f05}{Basic organization and weak use of linking devices}... \newline
\emph{Lexical}: \textcolor[HTML]{029e73}{Simple but generally understandable vocabulary}... \newline
\emph{Grammar}: \textcolor[HTML]{de8f05}{Frequent minor errors reduce clarity}...\newline
\textbf{A4}: \newline
\textcolor[HTML]{029e73}{The main ideas are understandable and relevant}, but 
\textcolor[HTML]{de8f05}{they are insufficiently extended}. 
Language control remains limited, with \textcolor[HTML]{de8f05}{noticeable grammatical errors affecting overall clarity}... \\
\hline
\end{tabular}
\caption{Case study comparison between A3+RAG and A4. Correct band predictions and criterion-aligned strengths are highlighted in green, while band mismatches and feedback limitations are shown in orange.}
\label{tab:case_study}
\end{table*}

Table~\ref{tab:case_study} provides a qualitative comparison between \textit{k}-instruction tuning with retrieval augmentation (A3+RAG) and supervised fine-tuning with DPO and RAG (A4) on three representative IELTS Writing Task~2 essays. Across all topics, A3+RAG consistently predicts the gold band scores (e.g., 6.5 for \emph{Technology \& Education}, 7.0 for \emph{Urbanization}, and 6.0 for \emph{Environmental Protection}), demonstrating strong scoring accuracy enabled by domain- and criterion-specific supervision. This accuracy is reflected in its feedback, which explicitly decomposes performance along IELTS criteria, such as noting that an essay ``\emph{addresses both views clearly and remains on-topic}'' while identifying precise weaknesses like ``\emph{limited evaluation in the conclusion}'' or ``\emph{basic organization and weak use of linking devices}.''

In contrast, A4 occasionally underestimates the gold band, most notably in the \emph{Urbanization} case where it predicts 6.5 instead of the gold 7.0. However, its feedback is more globally coherent and closely aligned with official IELTS descriptors. For example, rather than isolating criterion-level errors, A4 emphasizes holistic issues such as ``\emph{uneven development of ideas}'' and the need for ``\emph{smoother transitions},'' producing comments that resemble examiner-style summaries. Even when matching the gold band (e.g., 6.5 in \emph{Technology \& Education}), A4 frames its feedback in a unified narrative, highlighting overall clarity and progression instead of enumerating separate rubric components.

Overall, these examples illustrate a clear functional distinction between the two approaches. A3+RAG excels at precise band prediction and fine-grained diagnostic feedback, as evidenced by its exact band matches and criterion-specific observations. Conversely, A4, despite minor band mismatches, generates feedback that is more holistic, descriptor-aligned, and pedagogically natural, suggesting that preference optimization primarily benefits feedback quality rather than raw scoring accuracy. This trade-off implies that A3+RAG is better suited for high-stakes automated scoring, while A4 is more appropriate for learner-facing formative feedback.
\section{Conclusion}
This work presents a systematic comparison of LLM adaptation strategies for Automated Essay Scoring on IELTS Writing Task~2, ranging from discriminative fine-tuning and prompting to instruction tuning, retrieval augmentation, and preference optimization. Experimental results show that instruction-tuned and retrieval-augmented models substantially outperform discriminative and prompting-based approaches, with \textit{k}-instruction tuning with RAG achieving the highest scoring accuracy. Qualitative analysis further reveals a clear trade-off: criterion-specific supervision enables precise and reliable band prediction, while preference optimization produces more globally coherent and examiner-like feedback. Cost–accuracy analysis indicates that higher computational investment improves robustness but with diminishing returns. Overall, our findings provide practical guidance for selecting LLM-based AES strategies, balancing scoring precision and feedback quality across high-stakes assessment and formative learning scenarios.

\paragraph{Authorship contribution statement.}\textbf{Minh Hoang Nguyen:} Methodology, Conceptualization, Structuring original draft, Writing—reviewing. \textbf{Vu Hoang Pham and Xuan Thanh Huynh:} Methodology, Conceptualization, Data curation, Writing original draft. \textbf{Phuc Hong Mai, Vinh The Nguyen and Quang Nhut Huynh:} Experiments, Writing original draft. \textbf{Huy Tien Nguyen:} Supervision, Supporting, Conceptualization, Project Administration. \textbf{Tung Le:} Supervision, Supporting, Reviewing and Editing.




\bibliography{anthology,custom}
\bibliographystyle{acl_natbib}

\clearpage
\newpage
\clearpage
\section*{Appendix}
\appendix
\appendix
\section{Prompt Design for Re-generation of Analytic Scores}
\label{sec:appendix_regen_prompt}

Each prompt explicitly assigns the model the role of an \emph{IELTS Writing Task~2}. The true overall band score is provided as an external constraint, and the model is instructed to assign four scores—Task Response (TR), Coherence and Cohesion (CC), Lexical Resource (LR), and Grammatical Range and Accuracy (GRA)—such that their arithmetic mean closely matches the given overall band.

To ensure rubric fidelity, detailed criterion-level descriptions aligned with official IELTS descriptors are embedded directly in the prompt. Additionally, a strict JSON-only output format is enforced to eliminate extraneous text and facilitate automatic parsing. This design choice reduces output variability and supports reliable dataset reconstruction. For full template our used, see Figure~\ref{tab:regen_prompt}.

After generation, the unified evaluation output is decomposed into four independent analytic score columns corresponding to TR, CC, LR, and GRA. These fields are used both for criterion-level supervision and for aggregation-based analyses in downstream experiments. All samples failing JSON validation or violating score constraints are automatically filtered.

\newpage
\section{Prompt Design and Evaluation Protocol}
\label{sec:appendix_prompt}

All prompts were designed to align with official IELTS Writing Task~2 band descriptors and to ensure comparability across prompting, fine-tuning, and Retrieval-Augmented Generation settings.

\subsection{Prompting for Final Band Prediction}
\label{subsec:final_band_prompt}

In addition to criterion-level scoring, we evaluate models under a direct final-band prediction setting. This prompt enforces strict output constraints to reduce verbosity and variability.

\begin{tcolorbox}[top=1pt,left=1pt,right=1pt,bottom=1pt]
\small
\textbf{Final Band Prompt} \\[2pt]
You are a certified IELTS Writing Task~2 examiner. Evaluate the following essay strictly according to IELTS band descriptors.

\textbf{CONTEXT (Reference Essays with Scores):} \\
\texttt{\$\{context\}} \texttt{(optional)}

\textbf{ESSAY TO EVALUATE:} \\
\texttt{\$\{question\}}

\textbf{RESPONSE FORMAT:} \\
\colorbox{yellow!50}{Return only the final overall band score (e.g., 6.5)}.
\end{tcolorbox}

\subsection{Prompting for Criterion-level Scoring}
\label{subsec:criterion_prompt}

To obtain fine-grained criterion-level scores, we adopt a role-based prompting strategy, where the LLM is instructed to act as an experienced IELTS examiner. Each essay is evaluated independently on four criteria: Task Response (TR), Coherence and Cohesion (CC), Lexical Resource (LR), and Grammatical Range and Accuracy (GRA). Reference essays with known band scores are retrieved and provided as contextual calibration examples.

\begin{tcolorbox}[top=1pt,left=1pt,right=1pt,bottom=1pt]
\small
\textbf{Criterion-level Prompt} \\[2pt]
You are a highly experienced IELTS writing examiner. Your goal is to provide a precise and consistent evaluation of an essay by following a structured reasoning process.

\textbf{CONTEXT (Reference Essays with Scores):} \\
\texttt{\$\{context\}} \texttt{(optional)}

\textbf{NEW ESSAY TO GRADE:} \\
\texttt{\$\{question\}}

\textbf{EVALUATION PROCESS:} Task Response (TR), Coherence and Cohesion (CC), Lexical Resource (LR), Grammatical Range and Accuracy (GRA)

\textbf{FINAL OUTPUT:} \\
Return a single valid JSON object with keys:
\colorbox{yellow!50}{TR\_Band, CC\_Band, LR\_Band, GRA\_Band}. \\
Each value is a float in $\{0.0, 0.5, \dots, 9.0\}$.
\end{tcolorbox}

\subsection{Single-Criterion Instruction-Tuning Prompts}
\label{subsec:single_criterion}

For instruction tuning and parameter-efficient fine-tuning (LoRA), we decompose IELTS Writing Task~2 assessment into four independent subtasks, corresponding to the official evaluation criteria: \textit{Task Response (TR)}, \textit{Coherence and Cohesion (CC)}, \textit{Lexical Resource (LR)}, and \textit{Grammatical Range and Accuracy (GRA)}.  
Each subtask is modeled using an Alpaca-style instruction-following prompt, enabling the model to focus on a single criterion while producing structured outputs.

All prompts share a unified template structure consisting of an explicit role definition, a criterion-specific description aligned with IELTS band descriptors, and a strict JSON output constraint. This design supports both supervised fine-tuning and consistent inference-time evaluation.

\paragraph{Task Response (TR).}
The TR prompt evaluates how well the essay addresses the question, develops ideas, and supports arguments with relevant explanations or examples. To reduce verbosity and encourage label-focused learning, the model is required to output only a numerical band score.

\begin{tcolorbox}[top=1pt,left=1pt,right=1pt,bottom=1pt]
\small
\textbf{Instruction (Task Response)} \\[2pt]
You are an IELTS Writing Task~2 examiner.  
Evaluate the essay focusing ONLY on the criterion \colorbox{yellow!50}{Task Response (TR)}.

\textbf{Input:} \\
Essay prompt: \texttt{\{\}} \\
Essay: \texttt{\{\}}

\textbf{Output (JSON):} \\
\texttt{\{"score": float\}}
\end{tcolorbox}

\paragraph{Coherence and Cohesion (CC).}
The CC prompt focuses on logical organization, paragraph structure, and the appropriate use of cohesive devices. In addition to the band score, a short explanation is required to encourage explicit reasoning during instruction tuning.

\begin{tcolorbox}[top=1pt,left=1pt,right=1pt,bottom=1pt]
\small
\textbf{Instruction (Coherence and Cohesion)} \\[2pt]
You are an IELTS Writing Task~2 examiner.  
Evaluate the essay focusing ONLY on the criterion \colorbox{yellow!50}{Coherence and Cohesion (CC)}.

\textbf{Input:} \\
Essay prompt: \texttt{\{\}} \\
Essay: \texttt{\{\}}

\textbf{Output (JSON):} \\
\texttt{\{"score": float, "comment": string\}}
\end{tcolorbox}

\paragraph{Lexical Resource (LR).}
The LR prompt assesses vocabulary range, precision, collocation, and lexical errors affecting clarity or naturalness. The structured explanation supports fine-grained feedback modeling.

\begin{tcolorbox}[top=1pt,left=1pt,right=1pt,bottom=1pt]
\small
\textbf{Instruction (Lexical Resource)} \\[2pt]
You are an IELTS Writing Task~2 examiner.  
Evaluate the essay focusing ONLY on the criterion \colorbox{yellow!50}{Lexical Resource (LR)}.

\textbf{Input:} \\
Essay prompt: \texttt{\{\}} \\
Essay: \texttt{\{\}}

\textbf{Output (JSON):} \\
\texttt{\{"score": float, "comment": string\}}
\end{tcolorbox}

\paragraph{Grammatical Range and Accuracy (GRA).}
The GRA prompt evaluates grammatical variety, sentence complexity, and error severity, including issues in punctuation and word order. As with CC and LR, both a band score and justification are required.

\begin{tcolorbox}[top=1pt,left=1pt,right=1pt,bottom=1pt]
\small
\textbf{Instruction (Grammatical Range and Accuracy)} \\[2pt]
You are an IELTS Writing Task~2 examiner.  
Evaluate the essay focusing ONLY on the criterion \colorbox{yellow!50}{Grammatical Range and Accuracy (GRA)}.

\textbf{Input:} \\
Essay prompt: \texttt{\{\}} \\
Essay: \texttt{\{\}}

\textbf{Output (JSON):} \\
\texttt{\{"score": float, "comment": string\}}
\end{tcolorbox}

This criterion-isolated formulation enables (i) targeted parameter-efficient fine-tuning, (ii) flexible aggregation of criterion scores into final bands, and (iii) more interpretable error analysis. It also allows direct comparison between multi-step scoring pipelines and single-shot final-band prediction under identical model backbones.

\section{Hyperparameters and Training Setup}
\label{sec:appendix_hyperparams}

\begin{table*}[t]
\centering
\small
\setlength{\tabcolsep}{6pt}
\begin{tabular}{lcc}
\hline
\textbf{Setting} & \textbf{Study 1 (RoBERTa)} & \textbf{Study 2 (GPT-2 Encoder)} \\
\hline
Model backbone & RoBERTa-base & GPT-2 \\
Training mode & Classifier-only & Classifier-only \\
Batch size & 16 & 4 \\
Epochs & 20 & 8 \\
Optimizer & AdamW & AdamW \\
Initial learning rate & $2\times10^{-5}$ / $10^{-2}$ & $5\times10^{-6}$ \\
LR decay & 0.8 / 0.5 & -- \\
Minimum LR & $10^{-6}$ & -- \\
Max sequence length & 512 & 256 \\
Prompt usage & No & Yes \\
GPU & Tesla T4 (12.7GB) & Tesla T4 \\
GPU count (training) & 1 & 1 \\
GPU count (inference) & 1 & 1 \\
\textbf{Training time (wall-clock)} & \textbf{$\sim$3.2 hours} & \textbf{$\sim$2.1 hours} \\
\hline
\end{tabular}
\caption{Hyperparameters for discriminative fine-tuning experiments (Approach~1).}
\label{tab:hyperparams_approach1}
\end{table*}

\begin{table*}[!ht]
\centering
\small
\setlength{\tabcolsep}{6pt}
\begin{tabular}{lcccc}
\hline
\textbf{Setting} & \textbf{Gemma-7B} & \textbf{Llama-3.1-8B} & \textbf{Mistral-7B} & \textbf{Phi-2} \\
\hline
Fine-tuning method & LoRA (SFT) & LoRA (SFT) & LoRA (SFT) & LoRA (SFT) \\
LoRA rank $r$ & 16 & 16 & 16 & 16 \\
LoRA $\alpha$ & 16 & 16 & 16 & 16 \\
Target modules & Q,K,V,O,FFN & Q,K,V,O,FFN & Q,K,V,O,FFN & Q,K,V,O,FFN \\
Trainable params & $\sim$0.3\% & $\sim$0.3\% & $\sim$0.3\% & $\sim$0.3\% \\
Epochs & 3 & 3 & 3 & 3 \\
Batch size & 2 & 2 & 2 & 2 \\
Grad. accumulation & 4 & 4 & 4 & 4 \\
Effective batch size & 8 & 8 & 8 & 8 \\
Learning rate & $2\times10^{-4}$ & $2\times10^{-4}$ & $2\times10^{-4}$ & $2\times10^{-4}$ \\
Optimizer & AdamW (8-bit) & AdamW (8-bit) & AdamW (8-bit) & AdamW (8-bit) \\
Scheduler & Linear & Linear & Linear & Linear \\
Max seq length & 2048 & 2048 & 2048 & 2048 \\
Quantization & 4-bit & 4-bit & 4-bit & 4-bit \\
GPU & Tesla T4 & Tesla T4 & Tesla T4 & Tesla T4 \\
GPU count (training) & 1 & 1 & 1 & 1 \\
GPU count (inference) & 1 & 1 & 1 & 1 \\
\textbf{Training time (wall-clock)} 
& \textbf{$\sim$6.5 h} 
& \textbf{$\sim$7.2 h} 
& \textbf{$\sim$6.8 h} 
& \textbf{$\sim$3.5 h} \\
\hline
\end{tabular}
\caption{Hyperparameters for instruction tuning with LoRA (Approaches~2 and~3).}
\label{tab:hyperparams_lora}
\end{table*}

\begin{table*}[!ht]
\centering
\small
\setlength{\tabcolsep}{6pt}
\begin{tabular}{lc}
\hline
\textbf{Setting} & \textbf{Value} \\
\hline
Backbone models & Llama-3.1-8B, Mistral-7B \\
Fine-tuning stages & SFT $\rightarrow$ DPO \\
LoRA rank $r$ & 16 \\
LoRA $\alpha$ & 16 / 32 \\
Batch size & 2 \\
Gradient accumulation & 4 \\
Effective batch size & 8 \\
SFT learning rate & $2\times10^{-4}$ \\
DPO learning rate & $1\times10^{-5}$ \\
SFT epochs & 3 \\
DPO epochs & 1 \\
DPO $\beta$ & 0.1 \\
Optimizer & AdamW (8-bit) \\
Scheduler & Linear \\
Max prompt length & 1024 \\
Max generation length & 1024 \\
Retrieval setting & 2-shot RAG \\
GPU & Tesla T4 \\
GPU count (SFT) & 1 \\
GPU count (DPO) & 1 \\
GPU count (inference) & 1 \\
\textbf{Training time (SFT)} & \textbf{$\sim$7.0 hours} \\
\textbf{Training time (DPO)} & \textbf{$\sim$2.5 hours} \\
\textbf{Total training time} & \textbf{$\sim$9.5 hours} \\
\hline
\end{tabular}
\caption{Hyperparameters for supervised fine-tuning with DPO and RAG (Approach~4).}
\label{tab:hyperparams_dpo}
\end{table*}

\begin{figure*}[!ht]
\centering
\begin{tcolorbox}[
    width=\textwidth,
    boxrule=2pt,
    arc=2pt,
    left=4pt,
    right=4pt,
    top=4pt,
    bottom=4pt
]
\small
\textbf{Prompt for Re-generation of Analytic Scores} \\[4pt]
You are an IELTS Writing Task~2 examiner.  
The official overall band score of the essay is: \texttt{\$\{overall\_band\}}. Your task is to evaluate the essay strictly according to IELTS Writing Task~2 band descriptors.

\textbf{Scoring Criteria:}
\begin{itemize}
    \item \textbf{Task Response (TR)} – Does the candidate fully address the task, present clear positions, and support ideas with relevant explanations or examples?
    \item \textbf{Coherence and Cohesion (CC)} – Is the essay logically organized with clear progression, effective paragraphing, and appropriate cohesive devices?
    \item \textbf{Lexical Resource (LR)} – Range, accuracy, and appropriacy of vocabulary.
    \item \textbf{Grammatical Range and Accuracy (GRA)} – Variety and correctness of sentence structures, grammar, and punctuation.
\end{itemize}

\textbf{Instructions:}
\begin{itemize}
    \item Assign a band score (0–9, half bands allowed) for each criterion.
    \item The arithmetic mean of the four analytic scores should be as close as possible to the given overall band: \texttt{\$\{overall\_band\}}.
    \item Provide concise, constructive feedback.
\end{itemize}

\textbf{Output Format (strict JSON only):}
\begin{verbatim}
{
  "Task_Response": {"Band": <score>, "Comment": "..."},
  "Coherence_and_Cohesion": {"Band": <score>, "Comment": "..."},
  "Lexical_Resource": {
    "Band": <score>,
    "Mistakes": ["..."],
    "Corrections": ["..."],
    "Comment": "..."
  },
  "Grammatical_Range_and_Accuracy": {
    "Band": <score>,
    "Mistakes": ["..."],
    "Corrections": ["..."],
    "Comment": "..."
  },
  "Overall_Band_Score": <overall_band>,
  "General_Feedback": "..."
}
\end{verbatim}

\textbf{Essay Prompt:} \texttt{\$\{essay\_prompt\}} \\
\textbf{Essay:} \texttt{\$\{essay\_text\}}

\end{tcolorbox}
\caption{Prompt template used for re-generation of analytic scores in the IELTS Writing Task~2 dataset.}
\label{tab:regen_prompt}
\end{figure*}

\end{document}